\documentclass[twoside,11pt]{article}

\usepackage{jmlr2e}
\usepackage[utf8x]{inputenc}

\usepackage{amsmath}
\usepackage{graphicx}
\usepackage{amsmath}
\usepackage{amsfonts}
\usepackage{dsfont}
\usepackage{algorithm}
\usepackage[noend]{algpseudocode}

\makeatletter
\def\BState{\State\hskip-\ALG@thistlm}
\makeatother

\DeclareMathAlphabet{\mathcal}{OMS}{cmsy}{m}{n}

\DeclareMathOperator*{\argmax}{arg\,max}

\jmlrheading{1}{2018}{1-36}{10/18}{}{reisgonçalves}{João Reis and Gil Gonçalves}

\ShortHeadings{Hyper-Process Model: A Zero-Shot Learning algorithm for Regression Problems}{Reis and Gonçalves}
\firstpageno{1}

\begin{document}

\title{Hyper-Process Model: A Zero-Shot Learning algorithm for Regression Problems based on Shape Analysis}

\author{\name João Reis \email jpcreis@fe.up.pt \\
        \name Gil Gonçalves \email gil@fe.up.pt \\
       \addr SYSTEC, Research Center for Systems and Technologies\\
            Faculty of Engineering, University of Porto\\
            Rua Dr. Roberto Frias, 4200-465 Porto, Portugal
    }

\editor{}

\maketitle

\begin{abstract}

Zero-shot learning (ZSL) can be defined by correctly solving a task where no training data is available, based on previous acquired knowledge from different, but related tasks. So far, this area has mostly drawn the attention from computer vision community where a new unseen image needs to be correctly classified, assuming the target class was not used in the training procedure. Apart from image classification, only a couple of generic methods were proposed that are applicable to both classification and regression. These learn the relation among model coefficients so new ones can be predicted according to provided conditions. So far, up to our knowledge, no methods exist that are applicable only to regression, and take advantage from such setting. Therefore, the present work proposes a novel algorithm for regression problems that uses data drawn from trained models, instead of model coefficients. In this case, a shape analyses on the data is performed to create a statistical shape model and generate new shapes to train new models. The proposed algorithm is tested in a theoretical setting using the beta distribution where main problem to solve is to estimate a function that predicts curves, based on already learned different, but related ones.


\end{abstract}



\section{Introduction} \label{chap:intro}

The interpolation among different tasks and extrapolation of knowledge to new unseen tasks is a great challenge in the machine learning community and has been thoroughly explored in the past few decades. More specifically, one of the main areas that had brought significant advances in machine learning, such as the zero-shot learning (ZSL), is the transfer learning area. Contrary to the traditional machine learning setting, the main purpose of transfer learning is to reuse past experience and knowledge to solve the current problem. Normally, machine learning algorithms focus on isolated tasks that cannot be inherently used for other tasks. From all the training data associated with a specific task, the goal of transfer learning is to assist on the learning task for a future problem of interest. Therefore, this kind of domain proposes to solve the problem of transfer knowledge from different, but similar, tasks \citep{pan2010survey}. Techniques that enable knowledge transfer represent progress towards making machine learning as efficient as human learning \citep{torrey2009transfer}. This statement comes from the fact that humans do not learn from scratch all the new tasks they need to perform. Otherwise, the reuse of past experiences and acquired theoretical and empirical knowledge plays a significant role in how fast a human can learn.

The ZSL topic is framed into the transductive transfer learning setting, where it is considered unsupervised transductive transfer learning. It can be considered as unsupervised because no information (both input and output feature space) is used from the target task for learning, whereas most transductive solutions apply domain adaptation techniques between input feature spaces from both source and target tasks in order to learn a common feature space. In the context of transfer learning, source task is a problem already learned or solved and target task is the future problem to solve. This way, ZSL does not assume any input or output data from the target tasks, making the problem much harder to solve. However, some information should be provided in order to perform knowledge transfer from existing tasks (source tasks) to a new task (target task). This information is normally called side-information and is characterized by meta-information about the tasks themselves. It is based in this side-information that the relation among source and target tasks is learned. In sum, the requirements for ZSL are 1) already learned source tasks, e.g. as estimated functions, and 2) task descriptions from both source and target tasks. Hence, ZSL addresses a very specific problem from transfer learning that is very challenging to solve.

In the present work we propose a novel algorithm called hyper-process model (HPM) that differs from existing ZSL solutions by presenting a specific implementation for a regression setting. On one hand, most of the existing works in literature only address classification problems, like image or haptic data classification. On the other hand, there are a couple of works that generalize these algorithms to regression problems, but never propose a specific implementation that can truly leverage the properties of regression. This way, HPM performs a shape analysis to data and tries to correlate it with the task descriptions. From this perspective it would be possible to know what are the shape variations that most relate to certain task properties. The intuition behind using a shape analysis is that data itself have interesting properties to leverage that can be used to better understand task relations comparing with general frameworks. Additionally, these general ZSL frameworks make use of model coefficients to learn relations among source tasks, which is dependent on the method used for training that needs to be the same for all source tasks. By analyzing data drawn from trained models instead, this dependency is removed. In our perspective, this is key to improve the performance of ZSL problems for regression. Up to our knowledge, this is something that only has been explored in 2D and 3D computer vision settings, concretely in statistical shape models (SSM), where modes of deformation are learned to understand particular changes in shapes and ultimately allow to generate new shapes based on these modes of deformation. No algorithm in ZSL makes use of such shape analysis either for classification nor regression problems. 

The major contributions of the present work lie, first, in presenting a clear definition of a regression problem to ZSL. So far, only concrete definitions are available for classification problems and we truly believe that this regression definition to ZSL can help other researchers to easily frame their problems into this topic, and therefore discover already proposed works to assist solving their challenges. Secondly, we present an algorithm that is specific for regression problems in ZSL achieving better results than the ones proposed in general frameworks. Third, we explore the suitability of ZSL for regression in particular applications areas, such as chemistry and cyber security.

This work is organized in 4 more sections. Section \ref{chap_zero_shot} presents an extended related work and clear roadmap for ZSL area, where the most interesting works are detailed and discussed. Moreover, in Section \ref{chap_hyper_model} the proposed HPM algorithm is detailed along with all the methods used to build up the algorithm. In Section \ref{beta_distribution} a theoretical scenario is explored where the HPM is directly compared with an existing approach from general frameworks. Finally, Section \ref{discussion_concluion} presents an extensive discussion about the benefits of HPM over existing general frameworks, along with other application areas where ZSL can be successfully applied.

\section{Zero-Shot Learning}\label{chap_zero_shot}

One of the most intriguing and fascinating capabilities of humans is to generalize upon multiple and diverse tasks. When only presented with few examples, humans can quickly learn particular features of a certain object or task, distinguishing it from different classes of objects. The human capability to generalize allows to extrapolate and infer which kind of physical object it might be from previously seen examples of different object classes. As presented by \cite{biederman1987recognition}, humans have the capability to identify and distinguish about 30,000 objects, and for each of these objects there was no need for showing a million images of the same object in order to recognize and discriminate it from other objects, as it is often required in deep learning approaches such as in convolutional neural networks (CNNs). In fact, a great majority of humans would become confused if such an amount of images of the same object was shown to them. Instead, based on a small amount of images, or even from an object description, humans can generalize by extracting certain features of an object and form high level representations. By relating all the information learned and internal object representations, it is easier to learn from small amount of pictures. This capability to extract particular features and properties of an object and then generalize to other unseen classes of objects is one of the greatest challenges in artificial intelligence nowadays.

A definition for this particular type of problem was first presented by \cite{larochelle2008zero} where it first called zero-data learning and defines it as follows: \textit{"Zero-data learning corresponds to learning a problem for which no training data are available for some classes or tasks and only descriptions of these classes / tasks are given."}. Humans can imagine and mentally visualize certain objects when reading a book or an article, or just by thinking about certain past stories. Based on 1) a description of the object and 2) prior knowledge about the world, humans can materialize such imagined objects by drawing, sculpturing or even 3D modeling and recognize these if seen somewhere else. This is the main idea to explore in zero-data learning, that was afterwards named as zero-shot learning. If this description about an object is available, based on all the learning throughout lifetime humans can match their own mental visualization of an object with the physical one, and determine if these are the same or somehow similar in certain features. Normally, in such situations, intuition plays a significant role by matching an already learned object, problem or pattern, and immediately recognizing it without great effort. Such concepts are the ones that ZSL is based on to build a set of algorithms and strategies for machine learning.

The main motivation behind ZSL is that, as depicted and explored by \cite{larochelle2008zero}, the number of tasks is far too large and data for each task is far too little. We have already seen some great advances in artificial intelligence where systems reach superhuman capabilities in very specific tasks. Despite all these great achievements, these are not even close to the generalization of human capability and knowledge transfer from a set of tasks to new unseen ones. ZSL can be one of the tools to achieve such generalization capability. 


\subsection{Related Work} \label{related_work}

%
%

As already discussed, one of the first works related with the ZSL area is presented by \cite{larochelle2008zero} where the authors first make a definition of zero-data learning in order to distinguish their work from others and address specific issues that were not addressed until that time. In their work, they present two different approaches to the problem: 1) input space view and 2) model space view.

The first approach uses a concatenation of the input $x$ and the task / class description $d(z)$ for a given task $z$, and by using a supervised learning algorithm train a model $f^*(.)$ to predict $y_t^z$. Hence, for a new class $z^*$ and input $x^*$, one could predict the output by using $f^*([x^*,d(z^*)])$. The second approach is more model-driven, and is defined by $f_z(x) = g_{d(z)}(x)$. By defining a joint distribution $p(x,d(z))$ one can then set $g_{d(z)}(x) = p(x|d(z))$ and learn a probabilistic model that estimates the input $x$ belonging to class $d(z)$. However, a different way to achieve model space view is also presented. This uses the model parameters $\theta$ to train a model that maps class descriptions into model parameters. Assuming a family of functions $h_\theta(x)$ parametrized by $\theta$, if one defines a function $q(d(z))$ where the output is the same as the parameter space of $\theta$, then the output for a particular $x$ with class description of $d(z)$ is $h_{q(d(z))}(x)$. With this, the model space view is obtained by $f_z(x) = h_{q(d(z))}(x)$. 

For testing, 3 different datasets were used: 1) Character recognition in license plates, considering characters from 0 to 9 and A to Z and 3 others accentuated characters, summing up a total of 40 classes and 200 samples per class; 2) Handwritten alphanumeric character recognition with characters from 0 to 9 and A to Z with 39 examples per class; 3) Molecular compound dataset provided by a pharmaceutical company where the main idea is to develop a system that could identify if a molecular compound $x_t$ is active $y_t^z = 1$ in the presence of a biological agent $z$ for each of the 7 provided agents. The authors have used multiple machine learning techniques to model each of the problems, from support vector machines (SVM) to artificial neural networks (ANN).

The results for the character recognition show that the classification error tends to decrease as the number of classes increases, as expected, where the SVM with Gaussian kernel provides almost perfect discrimination of unseen characters. NNet-0-1 yields also good results being the second best technique. As for the handwritten characters all the models have nearly the same behavior by decreasing the error with an increasing number of classes. In both (1) and (2) datasets, there's no clear difference between input and model space view, apart from the SVM rbf that produced near perfect classifications. As for the molecular compound dataset (3), model space view performed better than input space view, performing better than random ranking.

%
%
Other interesting work that paved the way towards a more formal definition and theory of zero-shot learning is the work of \cite{palatucci2009zero}. In their work a two-stage approach is presented where the same concept as task / class description is used as before but now called semantic feature space. For their approach, the first stage is related with mapping brain images $X^d$ of dimension $d$ into a semantic feature space $F^p$ of dimension $p$, defining the following function $\mathcal{S} : X^d \rightarrow F^p$. Then, the second stage is to map this semantic feature space $F^p$ into the desired class label $Y$ where another function is defined $\mathcal{L} : F^p \rightarrow Y$. Hence, the main idea is to train a classifier $\mathcal{H}$ that can map the input $X^d$ into the correct class label $Y$ using the two presented functions, where $\mathcal{H} = \mathcal{L}(\mathcal{S}(.))$ called semantic output code (SOC) classifier. The main reason to separate the learning into two stages and avoid training directly a function is that one wants to predict class labels that are not present in the training phase. Therefore, the goal is to train $\mathcal{S}$ with a set of inputs that map into certain class labels, and train $\mathcal{L}$ with a larger spectrum of class labels.

The dataset used contains neural activity observed in 9 different human participants while watching 5 specific words from 12 different categories, summing up a total of 60 words. Two different knowledge bases were created for the semantic feature space for all 60 words, being one based on corpus5000 and the other on human218. As for the first stage, the authors used multiple output linear regression to learn $\mathcal{S}$ and 1-nearest neighbor classifier to learn $\mathcal{L}$. As for the experiments, $\mathcal{S}$ was only trained with 58 brain images, where a leave-two-out-cross-validation was performed, and $\mathcal{L}$ with all 60 image classes. This resulted in 3,540 comparisons and the approach achieved a performance of 80.9\% for the human218 and 69.7\% for the corpus5000. Moreover, the inputs for two classes left out of $\mathcal{S}$ were used representing a bear and a dog, and the classifier clearly distinguished between the two in almost all the 10 semantic questions selected for discrimination. Finally, the authors expanded the knowledge base used to train $\mathcal{L}$ using mri60 (with 60 nouns) and noun940 (940 nouns), and tried to predict the correct word for the held-out input and semantic feature using again both human218 and corpus5000. For this experiment, the authors calculated the median and mean rank accuracy, where for noun940 the median rank accuracy is above 90\% and mean about 80\% for human218 and for corpus5000 the median is around 79\% and mean of 70\%. As for the mri60, the median rank accuracy was around 88\% and mean 79\% for human218 and similar median and mean were obtained for corpus5000.

One of the key differences between the presented work and the one presented by \cite{larochelle2008zero} is that only the input of a brain image is required in order to classify the corresponding word for SOC, contrary to the need for both input and task description, as seen in the expression used by the authors $h_{q(d(z))}(x)$. This is one of the greatest advantages of using a two stage approach where classes can be learned in the latent space even when there's no input available for all the classes. However, one of the disadvantages is the training of two different functions where the performance of the function in the first stage greatly influences the performance of the whole classifier, even if the performance of the function in the second stage is good. This error accumulation of from the first to the second stage can invalidate the whole SOC approach for different application scenarios.

%
%
A similar two-stage approach called cross-model transfer (CMT) was proposed by \cite{socher2013zero} where the main idea is to train a model that is able to map image features into a word vector space, and then have a second model that is trained to classify these word vectors into the correct label classes. Again, it is assumed that more classes are present in the second stage rather than in the first. In the first stage, based on the work of \cite{coates2011importance}, the authors have extracted a set of unsupervised image features from raw image pixels in order to map these into a semantic space (word vector). Hence, for the semantic space, the authors have used an unsupervised model from \cite{huang2012improving} which is composed by 50-dimensional word space.

As for the second stage, the authors want to first assess if the presented image is from seen or unseen classes, so then labels can be chosen based on likelihood. The main motivation for such an approach comes from the analysis performed on the semantic features where images from unseen classes are close to related images from seen classes, but not as much as the images from the same seen class. One of the main goals of this approach is not only to develop a solution that yields good results for unseen classes, but also perform well in images that belong to already learned classes. For that, two novelty detection strategies were applied using outlier techniques. The authors have tested the approach in two different datasets, namely the CIFAR-10 and CIFAR-100. This represents one of the earliest works of ZSL and most of the recent ones do not use these datasets to test their implementation, so the results are not pertinent in this context, but only the technique itself.

%
%

Another interesting work worth referring that is also related with this two-stage approach was first introduced by \cite{lampert2009learning} and then further extended by \cite{lampert2014attribute}, where two different techniques were presented: 1) direct attribute prediction (DAP); and 2) indirect attribute prediction (IAP). The authors propose a probabilistic model as a way to address the problem of predicting the class labels of images, where the test classes were not seen / used in the training process. Hence, training classes $Y = \{y_1, ... , y_K\}$ are disjoint from $Z = \{z_1, ... , z_L\}$ test classes. For the DAP technique, a probabilistic model was used to estimate the probability of binary-value attributes given a certain image, so unseen images at test phase could also have an estimate into this attribute space. Hence, the authors modeled $p(a|x) = \prod_{m=1}^{M}p(a_m|x)$, where $a_m$ is the attribute representation and $x$ the corresponding image. Moreover, a probabilistic model was also trained to estimate the probability of a certain attribute set is from a specific unseen class. This model was defined as $p(z|a) = \frac{p(z)}{p(a^z)} [[a = a^z]]$, where $a^z$ is the attribute representation for the unseen class $z$, and $[[a = a^z]]$ is the Iverson's bracket notation \citep{knuth1992two}, where $[[P]] = 1$ if condition $P$ is true, and 0 is false. By combining both stages, the final probabilistic model is expressed as:

\begin{equation} \label{dap_prob} 
p(z|x) = \sum_{a \in \{0,1\}^M} p(z|a) p(a|x)
\end{equation}

The predictions from image to unseen class were then made using maximum a posteriori (MAP). As for the IAP technique, instead of a two-stage approach, an additional stage was used. First a mapping between image and training classes is performed, as a regular multiclass classifier, estimating $p(y_k | x)$ for each training class $y_k$. Then, a mapping between training classes and attributes is made $p(a_m | y) = [[a_m = a_m^y]]$, resulting in a model that maps images in attributes as $p(a_m | x) = \sum_{k=1}^{K} p(a_m | y_k) p(y_k | x)$. 

To test these techniques in a ZSL setting, the authors created the nowadays well famous to ZSL, animals with attributes (AwA) dataset, composed by over 30,000 images, 50 animal classes and 85 semantic attributes. Additionally, the authors also tested these in existing datasets such as the aPascal/aYahoo and SUN Attributes datasets. As for the experimental setup, using the AwA dataset, 40 classes were used in the training process and the remaining 10 in the test phase (5-fold cross validation). For the SUN Attributes dataset a 10-fold cross validation approach was used, meaning that approximately 637 classes were used for training and 70 classes for the test phase. 

The authors compare both DAP and IAP with two other methods from ZSL. Both have the same principle of training a classifier for the training set, and then estimate the most similar test class by using the trained classifier for the test images. This way, a test image is classified into a train class, and then the most similar test class is chosen using to different similarity criteria: 1) Hamming distance (CT-H) and 2) cross correlation (CT-cc). In the overall cases, the DAP and IAP are better than CT-cc and CT-H, where there is not much difference between both DAP and IAP.

%
%
The work of \cite{qiao2016joint} presents an algorithm that was greatly inspired by the DAP algorithm, where the authors explored the relations and dependence between the attributes to increase the performance of the system. For this purpose, the authors consider a chain of dependent attributes where the joint probability of each attribute for a specific class is calculated, contrary to DAP which calculates the marginal probability. However, due to high amount of attributes it is difficult to calculate these joint probabilities, so first a clustering algorithm is applied to organize attributes into sets. Only after this process these probabilities are calculated for each of the sets. Finally, the classes are predicted using MAP estimation, as used in DAP.

As for the datasets used, the authors have tested and compared their approach with the AwA and aPascal-aYahoo. Despite being an interesting work where different properties of attributes were explored, the results did not significantly improve compared with DAP. The best accuracy achieved in the AwA (aPascal-aYahoo) dataset is 44.14\% (24.4\%) while for DAP the accuracy was 42.5\% (22.6\%).

%
%

First introduced in \cite{akata2013label} and then extended and generalized by the same group \cite{akata2016label}, the attribute label embedding (ALE) is presented as an alternative that outperforms some of the DAP method limitations \citep{lampert2014attribute}. The authors state that ALE overcomes the limitations of being 1) a two-stage learning approach for ZSL problem, by 2) assuming the attributes on AwA are independent among themselves and 3) is not extendable to other sources of side information. Regarding 1) the problem is associated with not assuring that both attribute (first stage) and class prediction (second stage) are optimal because the learning process is not performed jointly, but separately. Hence, perhaps the prediction of attributes might by optimal, but not for class prediction. On 2) by assuming that attributes are independent, like "has stripes" and "has paws" for the AwA dataset, no additional information can be leveraged to increase the performance of the system, and they explore a hierarchical method to address such an issue. Finally, limitation 3) is related with only using the attributes available on AwA and no other complementary information such as textual descriptions that can be automatically processed. This last aspect is particularly interesting when little training data is available and other sources might increase the performance of the system.

Apart from the proposed ALE algorithm, the main contribution from the authors is a framework for learning label embedding with attributes in a ZSL problem. Hence, first a label embedding should be defined as a set of attribute vectors that correspond to a class label. The same way images can have some attributes that define stripes and color, the classes themselves also can have these attributes. Complementary to the previously described approaches, this means that both images and class labels have a latent representation of its own, instead of only the input image. To these latent representations, we should call image embedding to the image latent representation and label embedding to class latent representation. Hence, assuming that image embedding is defined by $\theta : X \rightarrow \tilde{X}$ and label embedding by $\varphi : Y \rightarrow \tilde{Y}$, the prediction function can be defined as such:

\begin{equation} \label{label_framework} 
F(x,y;W) = \theta(x)' W \varphi(y)
\end{equation}

where $W$ are the parameters that should be learned to predict the correct class $y$ for the input $x$. Based on this, the authors define an optimization problem to minimize the empirical risk and learn the model parameters in order to maximize the compatibility between image and label embeddings. Hence, the label embedding $\varphi(y)$ for each class can be learned as well from the data, the same way as $W$. The label embedding for all classes is defined as $\Phi$ and as a matrix of stacked $\varphi(y)$, where each row is a class. The only restriction is that the dimension of the embedding should be found, but a strategy such as cross-validation can be used. Another option is to define the label embedding a priori as side information, as normally occurs in the previous algorithms for the image embedding.

The authors define the ALE algorithm on top of the presented framework. For that, an already existing algorithm called web-scale annotation by image embedding (WSABIE) proposed by \cite{weston2010large} was used as a baseline to formulate ALE. For the optimization process, the authors use stochastic gradient descent (SGD) as a convex-function is not guaranteed. One of the greatest contributions of the this work is that ALE algorithm is compared with a handful of algorithms, apart from the DAP already stated. It is easy to see that if anyone is capable of defining a label embedding, the ALE algorithm can be readily used since these attributes are seen as side information. Therefore, the authors explore other kinds of embeddings such the hierarchical label embedding (HLE) first proposed by \cite{tsochantaridis2005large} or the word2vec label embedding (WLE) proposed by \cite{frome2013devise}. For the ZSL problem, these 4 algorithms were considered: DAP, ALE, HLE, WLE. Finally, the authors test the algorithms in the AwA dataset and CUB-200-2011 (CUB) with three different types of label embeddings: 1) Attributes that describe each class; 2) Hierarchical structure that represent each class; 3) Word2Vec based on English-language Wikipedia.

For the ZSL experiment, a 5-fold CV approach was used where 40 classes were used for training and 10 for testing. For the CUB dataset a 4-fold CV was used with 150 classes for training and 50 to test. The authors also test three different types of embeddings encoding: 1) Continuous between 0 and 1; 2) Binary being either 0 or 1; 3) Binary being either -1 or +1. The first assessment using only the ALE algorithm indicates a significant difference between continuous and binary encodings, favoring the continuous encoding. Also the regularized version of the optimization function was used where it seems that only the $\ell_2$-normalization benefits the performance and not the mean-centering $\mu$ parameter. As a direct comparison with DAP and ALE, ALE has a better performance with 48.5\% classification accuracy compared with 40.5\% from DAP.

As for the comparison between the three different proposed embeddings, ALE, HLE and WLE were compared in both datasets. In this experiment, ALE (AwA: 48.5\%; CUB: 26.9\%) perform better than HLE (AwA: 40.4\%; CUB: 18.5\%) and WLE (AwA: 32.5\%; CUB: 16.8\%) in both datasets. Additionally the authors tested the concatenation of ALE and HLE embeddings, where the best results were obtained with 49.4\% for AwA and 27.3\% for CUB.

%
%

The work presented by \cite{akata2015evaluation} proposes a new approach called structured joint embedding (SJE). The difference between SJE and the previously presented ALE algorithm is mainly on the optimization function, where the authors preferred the unregularized structured SVM as follows:

\begin{equation} \label{jse_compatibility} 
\frac{1}{N} \sum_{n=1}^{N} \argmax_{y \in \mathcal{Y}}\{0,\ell(x_n,y_n,y)\}
\end{equation}

where the loss function $\ell$ is the same as presented in ALE. For the optimization, again SGD was used and the regularization is performed by early stopping when using the validation set of cross-validation. The authors also present a additional approach based on multiple output embeddings. The algorithm learns the best transformation $W$ for a specific output embedding, and according to the given input embedding the best class is selected based on a confidence in each of the embeddings. As a certain output embedding can benefit more some classes than others, this approach uses multiple output embeddings and learns the best according to the provided input embedding. For this case, the authors, instead of using equation \ref{label_framework}, have updated the compatibility function as such:

\begin{equation} \label{jse_compatibility_multiple} 
F(x,y;W) = \sum_{k=1}^{K} \alpha_k \theta(x)' W_k \varphi_k(y)
\end{equation}

for $K$ output embeddings, and where $\sum_{k=1}^{K} \alpha_k = 1$. As for embeddings, the authors used input embeddings from a CNN presented in the DeViSe \citep{frome2013devise}, Fisher vectors (FV) used in ALE \citep{akata2013label} and features extracted from googlelenet (GOOG) \citep{szegedy2015going}. From the output embeddings, the features used were human engineered attributes, hierarchical features and text corpora also used in ALE. The datasets used were the AwA, CUB and standard dogs (Dogs), where Dogs do not have human annotated attributes.

One of the main conclusion is that the attributes engineered by man significantly increase the performance of the system, where for all datasets the combination of unsupervised and supervised embeddings performed better than unsupervised and supervised embeddings alone. This means that by including unsupervised extracted features can greatly benefit the ZSL setting.

%
%
The approach presented by \cite{xian2016latent} is called latent embeddings (LatEm), and is a direct extension of the SJE where a nonlinear piece-wise compatibility function is explored, opposed to the linear one used in SJE. This nonlinear compatibility is explored by learning a collection of linear models, where each linear model maximizes the compatibility among image-class embedding pairs. For the optimization routine, the same method as SJE is used where SGD is used. Hence, the authors present different approaches to optimize such a parameter and select the best model: 1) Find the best $K$ using a cross validation strategy by trying out 2, 4, 6, 8 and 10 linear models; and 2) Novel pruning based strategy. As the first approach is relatively straightforward, the intuition behind the pruning approach is that models that do not frequently maximize the compatibility between input and output embeddings are not of great importance and do not increase the performance, while increasing its complexity. The greatest benefit of the pruning approach when compared with cross validation is that only one model needs to be trained due to its adaptation during time to choose the best $K$ value.

The datasets used are the same as the ones used in SJE (AwA, CUB and Dogs) with both supervised and unsupervised embeddings so a direct comparison between the two approaches was made. By only using one type of embedding at a time and not making any sort of combination, the LatEm approach surpasses the SJE in all embeddings for all datasets, but for CUB with human annotated attributes. However, the best results are reported when the authors combine all the unsupervised embeddings with the supervised ones.

Some other interesting works were also proposed for the image classification problem in ZSL and worth mentioning, such as the deep visual-semantic embedding model \citep[see][]{frome2013devise}, the joint latent similarity embedding (JLSE) \citep[see][]{zhang2016zero}, the convex combination of semantic embeddings (CONSE) \citep[see][]{norouzi2013zero}, the semantic similarity embedding (SSE) \citep[see][]{zhang2015zero}, the embarrassingly simple approach to zero-shot learning (ESZSL) \citep[see][]{romera2015embarrassingly}, the synthesized classifiers (SYNC) \citep[see][]{changpinyo2016synthesized}, the semantic autoencoder for zero-shot learning (SAE) \citep[see][]{kodirov2017semantic}, the simple exponential family framework (GFZSL) \citep[see][]{verma2017simple}, the zero-shot classification with discriminative semantic representation learning (DSRL) \citep[see][]{ye2017zero}, the feature generating networks (FGN) \citep{xian2018feature} and the gaze embeddings (GE) for zero-shot image ilassification \citep{karessli2017gaze}. For a comprehensive survey of ZSL methods for image classification please refer to the work presented by \cite{xian2018zero}.

%
%

%
%

%
%

%
%

%
%

%
%

%
%

%
%

%
%

%
%

%
%
One of the most interesting applications of ZSL outside image classification domain is related with object identification using haptic devices presented by \cite{abderrahmane2018haptic}. In their work, the authors use the DAP algorithm to recognize a set of objects by grasping those with a robotic hand with tactile fingertips. The main idea behind the ZSL setting is to be able to correctly recognize an object that the system was not trained for. Hence, from cutaneous and kinesthetic information of the robotic hand, the system should correctly say that the object it is holding is, e.g. a plastic bottle, lamp or cup of tea, without any prior information about this specific object. For that, the authors used the attribute-based approach also presented in DAP, where, in this case, classes have associated a set of attributes that describe the object.

In order to test the proposed approach, the authors use a PHAC-2 dataset containing information about 60 different objects $Y$, with 24 annotated attributes $A$ describing those objects and both haptic $X_{b1}$ and kinesthetic $X_{b2}$ information. The results show a classification accuracy 39\% over a 5 random splits with 50 objects in the train set and 10 on the test set. Complementary to these results, the authors have set up a experiment with a real robotic hand with the same kind of sensing devices described before, where 20 objects were used together with 11 attributes that describe each object class. The tests were performed using three different methods: 1) local DAP (LDAP) where only one grasp was made; 2) data-fusion multi-grasp DAP (DF-MDAP) where multiple grasps were performed and "super-grasp" was calculated based on the mean values from all grasps; 3) similar classification for multi-grasp DAP (SC-MDAP) performs multiple grasps, and for each one gets a classification using LDAP until an object is classified with the same label $k$ times. The results show that SC-MDAP is the best approach for this setting, followed by SC-DAP and then LDAP, leading to almost 100\% accuracy in the test set with only 4 to 5 grasps.

%
%
In line with the previous work in the sense that task descriptors are used is the one presented by \cite{isele2016using} where the model parameters of a policy based approach in a reinforcement learning (RL) setting is predicted based on a set of defined task descriptors. This work makes use of the same principle as \cite{larochelle2008zero} and \cite{pollak2016models}, but the methods used to achieve it are different. The main goal of the present work is to jointly learn a sparse encoding of both model parameters $\theta^{(t)}$ from a policy $\pi_{\theta}$ and task descriptors $\phi(m^{(t)})$ in an latent representation, where $m^{(t)}$ is the task description for task $t$. Hence, in order to learn this sparse encoding the authors defined policy parameters as $\theta^{(t)} = L s^{(t)}$ and the encoding of task descriptions as $\phi(m^{(t)}) = D s^{(t)}$. Both $L$ and $D$ should be learned in order to reconstruct back the $\theta$ and $\phi(m^{(t)})$, where $s \in S$ should be a shared coefficient. To this joint learning the author call \textit{coupled dictionary learning} and to the whole algorithm task descriptors for lifelong learning (TaDeLL). The rational behind such algorithm is that similar task descriptions have similar policies, so information can be learned from these two different spaces. Therefore, the authors perform an adaptation to the policy gradient (PG), first introduced by \cite{sutton2000policy}, where both $L$ and $D$ parameters are optimized.

The authors perform a set of tests in three different simulated environments: 1) spring mass damper (SM); cart pole (CP); and 3) bicycle (BK). For each of the domain 40 different tasks were tested, where another 20 tasks were used to tune the regularizers values. The TaDeLL algorithm was compared with other algorithms, such as PG-ELLA, GO-MTL, single-task learning using PG. Additionally, the authors also tested a different version of TaDeLL called TaDeMTL where the learning is performed in a offline multi-task learning fashion. From the tests performed, TaDeLL performed the best in all the testing scenarios, representing a promising approach for the RL field, and more specifically for lifelong learning. Ultimately, the paper presents a general framework that is extensible to classification and regression problems by using their learning algorithm. As previously described, this kind of approach already explored first by \cite{larochelle2008zero} and around the same time by \cite{pollak2016models}, where a model of models was built making use of the model parameters applied to industrial scenarios.

\section{Hyper-Process Modeling}\label{chap_hyper_model}

So far, we have seen ZSL approaches that try to solve the problem of classifying new instances from classes that were not used in the training process. This means that a trained algorithm tries to correctly label a new instance from a class without being trained to do so. One of the key aspects to achieve a good performance is to have an additional feature space (often called latent space) that describes each task, where normally a meta-description of each task is used, apart from input and output feature spaces. This is one property that inspired the development of the proposed approach.

Despite the good results achieved in the works presented in the previous section, non of them are neither designed nor applied to regression problems. Regression maps certain inputs into a set of continuous output variables, while in classification the output is either 0 or 1, or in a range between 0 and 1 such as in probabilistic models. Nevertheless, regression is used in these works to help, e.g. map the inputs into a continuous latent space such as the coefficients of a linear classifier, as presented in \cite{larochelle2008zero}. However, the application of the problem is never regression. Most of the works are related with classification problems ranging from image classification to molecular compound matching or object classification from haptic data. Here, we present an approach called hyper-process model (HPM) that addresses the problem of ZSL for regression problems. Although, it should be stated that some of these works presented in Section \ref{chap_zero_shot} are general enough to be used in regression applications, but were not designed to take advantage on its inherent properties, such as output continuous variables.

For the remaining of this Section, we will first make a definition of the ZSL problem in a regression setting, and then describe all the methods used to build up the HPM. Finally, we will present and describe the proposed algorithm.

\subsection{Problem Definition}

As a first step, we would like to define the problem of ZSL to regression. Up to our knowledge, this is the first work that makes such a definition for regression. Related with image classification, most of the ZSL techniques take advantage on the difference between input images, which is something normal where two different objects are displayed. Assuming inputs for a certain class / task as $X_i \in \mathcal{X}$ for class $i$, we can say that these techniques assume $P(X_i) \neq P(X_j)$ where marginal distributions among classes are not the same. This means that the difference between the images can be learned to separate both from different classes. Contrary to this, for ZSL in regression problems the inputs for different tasks could be the same and the responses might be different according to their specific task. For example, the amount of traffic in different parts of the city can be the same $P(X_i) = P(X_j)$, where $i$ and $j$ represent different parts of the city, but the air quality might be different because of different amounts of vegetation. If one part of the city has more vegetation, the air quality is higher, and vice versa. Most of the works first try to map the input into a latent space, which normally is a task descriptor, that can be generally expressed as $\mathcal{G} : X^n \rightarrow F^k$, where $X \in \mathbb{R}^n$ are inputs and $F \in \mathbb{R}^k$ are the task descriptors. In order to successfully learn the differences between tasks or classes, there should exist some difference between the task inputs like cubes and spheres, or cats and houses. Therefore, the assumption of $P(X_i) \neq P(X_j)$ is implicit in the context of image classification, which might not hold true for regression. Hence, this draws the first difference between ZSL works for classification and regression, where it is not assumed that the marginal distribution of inputs from different tasks is different, and hence the proposed technique is applicable for problems where $P(X_i) = P(X_j)$.

Additionally, another key difference between ZSL for classification and regression is that multiple image classes are learned at the same time as a multi-task learning fashion. For the particular case of ZSL in classification, we have already seen from Section \ref{chap_zero_shot} that this learning normally occurs in two different steps: 1) Learning a mapping between inputs and task description, and 2) Task description into class labels. This means that only one classifier should learn the differences between images and correctly predict the corresponding task description, and also a classifier that handles the predicted task description and correctly classifies it into the desired class labels. Ultimately, the final goal of ZSL in classification is to provide a new unseen image and correctly predict the label from a class not used in the learning process. Opposite to this idea, for the regression setting, the main idea is to build a whole new predictive function suitable for the new unseen task, where multiple inputs can be fed as a regular regressor. Therefore, for each source task, a regressor needs to be previously learned and together with the task description, a new function should be derived for a target task. The only work that uses the same approach is the already depicted technique called model space view presented by \cite{larochelle2008zero}. Additionally, the same principle was applied to solve a concrete problem in the area of manufacturing systems named hyper-model (HM) \citep{pollak2016models}. Despite these techniques being in fact applicable for regression, the proposed approach overtakes some limitations of such techniques. These will be presented later in this section, and will be highlighted and explained with a theoretical example.

In sum, we can define the ZSL for regression problem in the context of this work as the generation of a predictor that can be used in a new, unseen task, based on 1) task descriptions for both source and target tasks and 2) a set of predictors, one for each source task. Hence, we should define a task description as $c_i \in C$  for task $i$, where $C$ is defined as all the source task descriptions; and the predictors as $f_i \in F$, where $F$ is defined as a set of functions. For the latter, we should define a function as $f_i : X \rightarrow Y$, where $X$ and $Y$ are the input and output feature spaces, correspondingly. Additionally to all of this, we should also define a function that maps the task descriptions into a latent space $\mathcal{L} : C \rightarrow Z^p$, where $Z$ is the latent space in a $p$-dimensional space. If each of the predictors of the source tasks has a set of trained parameters $\theta$ and all the predictors have the same number of parameters, this approach would be identical to the one presented by \cite{larochelle2008zero} where $Z^p$ represents the same as $\theta$, so the parameters of the new function $\theta'$ would be predicted by $\mathcal{L}$ providing the target task description $c_t$, being $t$ the target task. However, the key difference between the proposed approach and the one presented by \cite{larochelle2008zero} is that the feature space $Z^p$ is not the a set of function coefficients. In the proposed approach the feature space is independent from the function coefficients and in fact do not assume that the number of coefficients should be same for all the functions used to learn the source tasks. For example, in order to learn a predictor that maps task descriptions into function coefficients, one should choose the type of machine learning technique to use, such as degree 2 polynomial, to train all the source tasks. In \cite{larochelle2008zero} and \cite{pollak2016models}, this implies that all source tasks will be trained using the same technique, not exploring the possibility of using the best machine learning technique for each source task. We interpret this as a limitation, where different tasks might have different complexities, and therefore certain types of functions might be more suitable to some tasks, and not to others. In the proposed approach we make use of a widely known technique from the computer vision area to address such a limitation, and create a common feature space for different machine learning techniques.

\begin{figure}[!ht]
\vskip 0.2in
\begin{center}
\centerline{\includegraphics[width=6in]{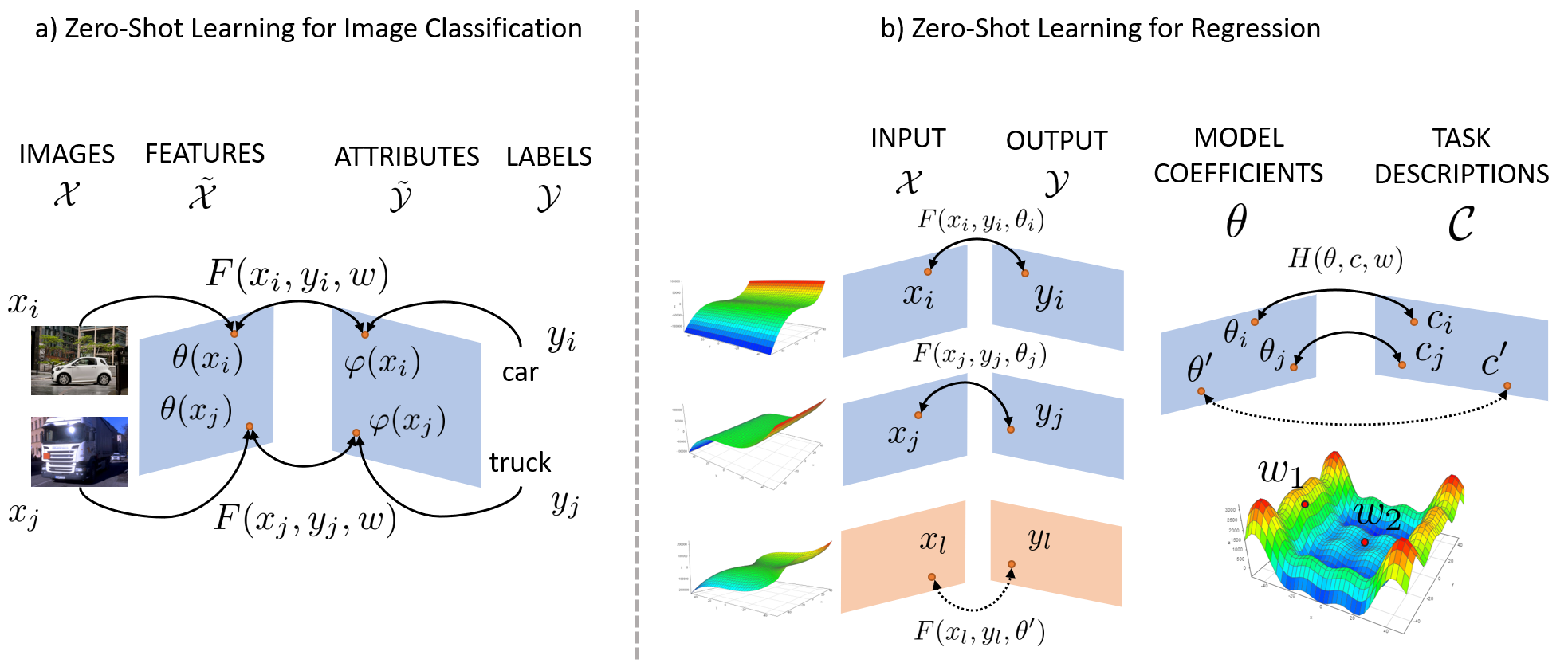}}
\caption{Comparison between ZSL for image classification and regression. a) Case where a latent representation for both images and classes is used, and a compatibility between these is learned \citep{akata2015evaluation}. b) Case where multiple models are used to learn a hyper-model that maps model coefficients $\mathcal{\theta}$ into task descriptions $\mathcal{C}$. Upon new task descriptions $c'$, new model coefficients $\theta'$ can be estimated and a new model is created \citep{pollak2016models}. This representation is also applicable in the model space view approach from \cite{larochelle2008zero}.}
\label{class_vs_regre}
\end{center}
\vskip -0.2in
\end{figure}

For a more complete explanation, Figure \ref{class_vs_regre} makes a visual comparison between two approaches as a way to clearly make a distinction of ZSL for regression from ZSL for classification, in particular, to image classification. This way, on the left-hand side is a representation of the SJE approach \citep{akata2015evaluation} that makes use of two latent spaces, namely image embeddings and class embeddings as presented in Section \ref{related_work}. In this setting, the main idea is to present an unseen image from an unseen class during training, and correctly estimate its label. Contrary to this, the goal of ZSL for regression is to estimate a new model by making use of an unseen task description and previous knowledge about already existing models. Particularly for the hyper-model approach this learning is simply the mapping between coefficients and task descriptions of source models used to estimate the target model. This way, on the right-hand side two stages can be clearly seen. One is related with training the source models to derive the best models' coefficients $\theta$ for each task and the other is to train the hyper-model using those coefficients and existing task descriptions. Once a new task description is available, the model coefficients $\theta'$ can be estimated and a new function can be used $F(x_l,y_l,\theta')$, where $x_l$ are the new input values that can be used to predict $y_l$ (orange boxes on the bottom represent the new generated function). Hence, this visual separation allows to clear draw the main differences between classification and regression settings for ZSL, where one tries to label unseen instances in a class not used during training, and the other tries to estimate a whole new function based on previous acquired knowledge of existing functions and task descriptions.

In the next two subsections, we will be presenting two different methods used to build the HPM. First we will introduce the hyper-model concept proposed by \cite{pollak2016models} for process models in manufacturing applications, and secondly present the statistical shape model (SSM) first proposed by \cite{cootes1995active} for image segmentation. Ultimately, the HPM can be viewed as an extension to the hyper-model it self, and hence its name.

\subsection{Hyper-Model} \label{hyper_model_sec}

The hyper-model concept was introduced by \cite{pollak2016models} where a model of models is built and applied to industrial scenarios. Complementarily, the authors introduce the notion of \textit{condition} that are fixed quantities that govern a certain industrial process, like thickness in metal sheets for welding processes or deep drawing. This concept of condition is what defines each task in the context of ZSL, where different conditions mean different tasks. Assuming that a model is a set of base functions that transforms a certain input into an output, a model has always associated a condition that quantitatively describes the task to learn. Based on this, the main idea is to build a hyper-model to generate models for a whole continuum of conditions, aiming at mapping model coefficients from the base functions into a set of conditions. This way, by providing a set of new conditions, it is possible to derive a new set of coefficients and build a new model for prediction in the context of those conditions only. As one might have realized by now, this approach is independent from being a regression or classification problem. As far as the coefficients of base functions and conditions are available, the hyper-model can be applied to both settings.

Defining these conditions as $\boldsymbol{\varsigma}_n \in \mathbb{R}^c$ for the $n$\textsuperscript{th} model and $c$ feature vectors, and a model as $\boldsymbol{z}_n = f_{\lambda_n}(\boldsymbol{x})$ that map an input $\boldsymbol{x}$ into an output $\boldsymbol{z}_n$, being $\lambda_n$ the process model coefficients that build $f_{\lambda_n}$, is possible to create a hyper-model. Thus, the model can be represented as a linear combination of some base functions $\boldsymbol{\phi}$ and the process models coefficients $\lambda_i$

\begin{equation} \label{process_model} f_{\lambda_n}(\boldsymbol{x}) =  \sum_{i=1}^{N} \lambda_i \boldsymbol{\phi}_i(\boldsymbol{x}) \end{equation}

Based on this, the hyper-model allows to relate the model parameters with conditions represented by the following expression:

\begin{equation}\label{hyper_model} \boldsymbol{\varsigma} =  g_{\boldsymbol{\beta}} (\boldsymbol{\lambda}) \end{equation}

where $\boldsymbol{\beta}$ are the hyper-model coefficients. Ultimately, also the hyper-model can be expressed as a linear combination of some base functions $\boldsymbol{\Psi}$ and the hyper-model coefficients:

\begin{equation} \label{hyper_model_linear} g_{\boldsymbol{\beta}} (\boldsymbol{\lambda}) = \sum_{k} \beta_k \boldsymbol{\Psi}_k (\boldsymbol{\lambda}) \end{equation}

This way, we formulate the problem as finding the hyper-model coefficients to derive a transformation function that maps the model coefficients $\boldsymbol{\lambda}$ to conditions $\boldsymbol{\varsigma}$. However, as the authors state, a necessary condition for the previous formulation of a hyper-model is a homogeneous representation of all involved task functions, meaning that the base functions used for the models need to be the same, like using a degree 2 polynomial for all the models. This means that if we want to use different base functions for different tasks we need to find a way to bring the model coefficients into the same common representation.

The method proposed to tackle this limitation lies in the idea of not using explicitly the coefficients $\boldsymbol{\lambda}$ of the models, but instead, take a step back and directly use data. As the most suitable machine learning techniques to use highly depend upon data, if two different datasets have different properties, different techniques might be applied, like a support vector regression in one dataset and linear regression in other. As stated by \cite{wolpert1997no} from the no free lunch theorem, "\textit{if an algorithm performs well on a certain class of problems then it necessarily pays for that with degraded performance on the set of all remaining problems}". Hence, if different datasets have different types of complexities and properties, the same technique will perform well in some of these datasets, and worse in others.

In this context, the optimal solution for such a problem is to use the best algorithm possible for each dataset and takes advantage on this to build the hyper-model. This means that we should use a technique that makes different techniques comparable among themselves and bring them to the same level of abstraction. To that intent, our proposal is to use directly the data instead of the model parameters. In the next subsection we will detail the SSM approach that is suitable to explore the properties of data and therefore, a suitable candidate, but not the only one, to achieve this common representation to train the hyper-model.

\subsection{Statistical Shape Modeling} \label{statistical_shape_model}

The statistical shape model (SSM) \citep{cootes1995active} is a widely used technique for image segmentation that analyzes the geometrical properties of a set of given shapes or objects by creating deformable models using statistical information. As a mathematical transform to be applied to these set of shapes, the most common techniques are principal component analysis, approximated principal geodesic analysis, hierarchical regional PCA \citep{mesejo2016survey} and singular value decomposition, where non-affine modes of deformation are calculated.

The problem definition for this area of research is framed as a maximization problem to overlap a deformable model in the object / region of interest. The overall idea behind this method is to obtain the optimal affine (rotation, scale and position) and non-affine transformation parameters where a deformable model is built and matches a segment of an image. The same way this method assumes that there exist specific shape variations and these can be quantified forming a deformable model, is the same way that we assumed that these variations also exist in different tasks, and a deformable model for a set of tasks can be derived.

There exist multiple recent examples that use such an approach, most commonly in medical imaging, like \cite{Shakeri201658} that proposes a novel approach of groupwise shape analysis that is able to analyze two study groups (healthy and pathological). The main idea is perform a morphological study to predict neurodevelopmental and neurodegenerative diseases, such as Alzheimer's, by quantifying sub-cortical shape variations by using statistical shape analysis. Another example is presented by \cite{nguyen2016use}, where the authors used the SPHARM-PDM (SPherical HARMonic) framework, introduced by \cite{styner2006framework}, to analyze 20 patients undergoing bilateral sagittal split osteotomy. In another work presented by \cite{Shin2016252} the authors used statistical shape analysis to optimally obtain the landmarks for midsagittal reference plane for evaluation of facial asymmetry. Moreover, in \cite{Yates20161540} the authors study the statistical mean and boundary models of the human spleen in an occupant posture, by using PCA to find the modes of variations and boundary models.

In a more formal way, a shape can be defined as \textit{"all the geometrical information that remains when location, scale and rotational effects are filtered out from an object"} \citep{dryden1998statistical}. Therefore, the information about a shape can be represented in different formats such as landmarks, parametric description and deformation-based \citep{Zhang2016155}. On one hand, landmarks are a set of points used to describe the shape reliably, where methods can be used to calculate these automatically or be manually annotated, e.g. in a CT scan. On the other hand, the parametric descriptors are functional approximations of the shape, and therefore limiting shapes to a set of coefficients. Ultimately, the deformation representation is based on shape matching between images and a template, considering smooth constraints in the deformation field.

Defined as \textit{a point of correspondence on each object that matches between and within populations} \citep{dryden1998statistical}, a landmark is a point on a shape that as a direct correspondence in all the other shapes used to build a deformable model. Is this correspondence between points that allows the statistical processing to calculate the deformation of each shape in relation to its mean. Normally, the landmarks are described as a $kn$ element vector $\boldsymbol{x}$, where

\begin{equation} \boldsymbol{x} = [x_1, x_2, ... , x_n, y_1, y_2, ... , y_n]^T \end{equation}

in this case, for the dimensionality of the landmark representation space we have $k$=2, and $n$ the number of landmarks of a given shape. Hence, to create a deformable model we need multiple shapes

\begin{equation} \boldsymbol{x}_i \in \mathbb{R}^{k}, i = 1,...,N \end{equation}

being $N$ the number shapes available for analysis. Here we consider a shape as belonging to a specific model or task, and the landmarks of each shape are all the instances from a dataset, where both input and output features can be included. Therefore, each shape is the dataset to build a model. However, since we cannot ensure beforehand that all the collected datasets have the same size for each task, we also cannot assume that the datasets used to build the models are suitable to create the deformable model (remember that all the shapes need to have the same number of landmarks and they should match between each others). Instead, we can take advantage on the generalization capability of the models and sample a dataset per model, which will be considered a shape. This will guarantee that all the shapes have the same number of landmarks. For this intent, we need to ensure that the inputs provided to all the models are the same to guarantee the consistency of all shapes. Again, remember the assumption in the context of the present work of $P(X_i) = P(X_j)$ where the distribution of the input feature space might be the same for all models, and hence is valid to use the same values of inputs to draw a new dataset from a model. Hence, to build the deformable model, only the output data should be used since no information is gained from using input data. Nevertheless, the proposed HPM algorithm can be generalized to regression problems with $P(X_i) \neq P(X_j)$, where input data can be included as far as the landmarks match between each other. 

In order to build the deformable model, a mathematical transformation needs to be learned. One of the common techniques used to learn this transformation is the principal component analysis (PCA) introduced by \cite{Flury1988} which assumes a multi-variate Gaussian distribution. We will be describing the decomposition process using PCA, which is composed by the following steps:

\begin{enumerate}
\item Calculate the mean shape: \begin{equation} \label{mean_equation} \bar{ \boldsymbol{x}} = \frac{1}{N} \sum_{i=1}^{N} \boldsymbol{x}_i \end{equation} where $N$ is the number of shapes;

\item Calculate the covariance matrix: \begin{equation} C = \frac{1}{N} \sum_{i=1}^{N} DD^T \end{equation}

where \begin{equation} D = ((\boldsymbol{x_1} - \bar{ \boldsymbol{x}}), ... , (\boldsymbol{x_N}  - \bar{ \boldsymbol{x}})) \end{equation}

\item Calculate eigenvalues and eigenvectors of C: \begin{equation} C \phi_k = \lambda_k \phi_k  \end{equation}

where $ \boldsymbol{\phi}$ are the eigenvectors and $ \boldsymbol{\lambda}$ are the eigenvalues.
  
\end{enumerate}

When there are fewer instances in the dataset when compared to the number of dimensions, the eigenvectors and eigenvalues can be efficiently calculated as follows:

\begin{equation} \frac{1}{N} D^T D q_k = \mu_k q_k \end{equation}

multiplying both sides by $D$ we obtain

\begin{equation} \label{eq_eigen} C(D q_k) = \mu_k(D q_k) \end{equation}

From equation \ref{eq_eigen} we infer that $\phi_k = D q_k$ and $\lambda_k = \mu_k $. The result of PCA is a decreasing order of the non-negative eigenvalues that represent the significance of each eigenvector (principal component). These eigenvectors are the modes of variations that allow to deform the model. With the most significant eigenvectors calculated, we can now approximate any training set, $\boldsymbol{x}'$, using the following expression:

\begin{equation} \label{deformable_model} 
\boldsymbol{x}' = \bar{ \boldsymbol{x}} +  \boldsymbol{\phi}  \boldsymbol{b} 
\end{equation}

where $\boldsymbol{b}$ are the parameters for the deformable model, and is given by the following expression:

\begin{equation} \label{process_parameter}  \boldsymbol{b} = \boldsymbol{\phi}^T ( \boldsymbol{x} - \bar{ \boldsymbol{x}}) \end{equation}

The variation of $\boldsymbol{b}$ allows to change the shape of the deformable model, and normally they are constrained to $ \pm 3 \sqrt{\lambda_i} $ to provide similar shapes to those present in the original dataset.

Based on this, the common representation to be used in the hyper-model are the $\boldsymbol{b}$ values that allows to reconstruct the original shape, instead of using explicitly the model coefficients. We consider the use of the statistical shape model concept the key to explore the best out of the each dataset properties and complexity. The approach taken to learn the deformable parameters for each shape can be seen as an unsupervised way to create a common space where multiple tasks have the same representation. If one uses different techniques to model the datasets as a way to increase generalization, the most organic step to take is to find a way to translate the different representations of model coefficients into the same common space. Despite existing different approaches to create a common feature space from different models in an unsupervised way, taking the step to generate data from the trained models to form shapes, build a deformable model and ultimately use the parameters of this deformable model seemed the most effective, and above all, flexible way of creating this common representation. 

In the following subsection the proposed algorithm will be detailed, which will glue together the presented methods of hyper-model and statistical shape model.

\subsection{Proposed Approach} \label{proposed_approach}

The main intentions of the present section is to, first, clearly present the full algorithm of hyper-process model (HPM) from the point of using the models trained with different techniques, to the final estimation of the new model to be used as a predictor in a new task. Secondly, it is intended to be reproducible for other researchers, where a step by step description of the algorithm is presented and explained. For that, most of the equations, notations and notions presented earlier are used, being the algorithm description just an organized way to present the approach.

Hence, Algorithm \ref{zsl_algorithm} presents all the steps required to implement the solution for different contexts of application. The first thing to notice is that the algorithm itself is divided into two different parts, as in the previous two subsections. This was intended so readers can easily relate to what was explained before and quickly find the content associated to each technique. Based on this, the algorithm starts to introduce all the parameters necessary for its execution. As described, all the trained models are required along with the corresponding conditions (which are the task descriptions from ZSL). Moreover, the target condition is required in order to generate the new model. Additionally, one should also specify the number of landmarks to use for each shape, together with two more vectors that define the minimum and maximum values for the input features space. These minimum and maximum vectors are required so one could generate the input values to sample from the trained models. Since we are assuming $P(X_s) = P(X_t)$, only a vector is required and is used in all source tasks to generate shapes. Finally, we assume to have $m$ trained models to deal with.

For this algorithm, the SSM first comes into place because the hyper-model is dependent on the common representation of models to be trained. Hence, the first step (line 3) is to generate the input values $X$ according to the minimum, maximum and number of intended landmarks per shape. Since we assume that no information can be drawn between the different inputs from the various models (as stated by $P(X_i) = P(X_j)$), the same input values are used for all the models. Therefore, the shapes $S_i$ are built only considering the values from the output feature space, as presented in line 5, where $i$ is a specific model.

\begin{algorithm}[h!]
\caption{Hyper-Process Modeling} \label{zsl_algorithm}
\begin{algorithmic}[1]
\Procedure{HPM($F,\varsigma,\varsigma',n,min,max$)}{$F$ is a set of source models, $\varsigma$ is a set of conditions associated with each source model, $\varsigma'$ is the target condition to be used for model generation, $n$ is the number of data points per shape, $min$ and $max$ are vectors of size $r$ (assuming $	 X_i \in \mathbb{R}^r$) with minimum and maximum values for the input features, correspondingly. Finally, $m$ in the number of source models.}
\BState \emph{Statistical Shape Model}:
\State Define the input to sample from existing models: $X \gets GenerateInput(min,max,n)$
\For {$i = 1 \rightarrow m$}
\State Get shape: $S_i = f_i(X)$
\EndFor
\State Get the mean shape: $\bar{S} = \frac{1}{N} \sum_{i=1}^{N} S_i$
\State Get eigenvectors from PCA decomposition: $\phi \gets PCA(S)$
\State Get deformable parameters from PDM: $\boldsymbol{b} = \boldsymbol{\phi}^T ( S - \bar{S})$
\BState \emph{Hyper-Model}:
\State Train the hyper-model: $h : b \rightarrow \varsigma$
\State Get the deformable parameter for new shape: $b' = h^{-1}(\varsigma')$
\State Get new shape: $S' = \bar{ \boldsymbol{S}} +  \boldsymbol{\phi}  b'$
\State Train a model for the new task. $f' : X \rightarrow S'$

\State \textbf{return} $f'$.

\EndProcedure
\end{algorithmic}
\end{algorithm}

The next step is calculate the mean shape from all the generated shapes (line 6), where equation \ref{mean_equation} is used. In order to get all the eigenvectors to build the deformable model, a decomposition needs to be performed on all the generated shapes and PCA is applied (line 7). One should emphasize again that each shape is a vector of $kn$ elements, where $k$ is the number of features and $n$ is the number of landmarks to use. Therefore, PCA is performed on a $m \times kn$ matrix $S$ composed by all the shapes from source models, where these shapes are stacked in rows. Finally, the last step for the SSM is to derive all the deformable parameters for all the models (line 8). This way, equation \ref{process_parameter} is used. These are the parameters required to generate back the initial shape based solely on the deformable model. In order to get a good shape reconstruction the number of components chosen when performing PCA is critical, being a trade-off between reconstruction and complexity. On one hand, if few components are chosen, the greater the reconstruction error will be but less dimensions are required, and thus, less complex the problem is. On the other hand, if all the components are chosen, the reconstruction error will be minimum, but the complexity of the problem is far to great to deal with. In these situations, a good rule of thumb is to use the number of components (ordered by decreasing order of model variance) that attend for a cumulative sum of variance of at least 95\%.

After building the deformable model, together with all the deformable parameters, the hyper-model is ready to be trained. For this case, and as presented by \cite{pollak2016models}, one should train a hyper-model using any machine learning technique that seems suitable for the problem, by mapping deformable parameters into conditions. One might think at this stage that would be more suitable to map conditions into deformable parameters instead, because we can use the trained model to predict the parameters based on new conditions. However, in most of the cases the dimension of the deformable parameters are greater than conditions, so the modeling needs to be made according to line 10. Only in the cases where 1) the dimension of parameters is the same or lower than conditions or 2) multiple models are trained as a hyper-model an each one of those models has only an output variable different from the others, the model can be trained as follows $h : \varsigma \rightarrow b$. The implication of building a hyper-model that maps deformable parameters into conditions is visible in line 11, where the technique used needs to be invertible in order to get the new deformable parameters according to the specified new conditions. As an alternative, the level set where the model surface intercepts with the hyper-plane for the intended target condition can be calculated, as performed in the work of \cite{pollak2011retrieval}, or formulate a minimization problem where the distance between the predicted and target conditions should be minimized. Once the deformable parameters are obtained from the hyper-model according to the target conditions, the next step is to generate a new shape based on equation \ref{deformable_model} as presented in line 12. The last step is to train a model to map the initially generated input values into the generated shape, which corresponds to the output values for that specific condition. 

Although being out of the scope of the present work, we would like to introduce a new version of the HPM algorithm where $P(X_i) \neq P(X_j)$, detailed in Algorithm \ref{zsl_algorithm2}. From this assumption, we could not only learn new information about the various output feature spaces from different tasks, but also learn about the input feature spaces. The only restriction about this approach is that the input feature space among different tasks should be the same $X_i = X_j$ where different distributions can be assumed. We consider this algorithm an expansion on the previous to a more general a broad application. Hence, we call this algorithm HPM2, not only for being the second version of the algorithm but also because it contemplates the two input and output feature spaces in the context of ZSL.

\begin{algorithm}[h!]
\caption{Hyper-Process Modeling - Extension with input feature space}\label{zsl_algorithm2}
\begin{algorithmic}[1]
\Procedure{HPM2($F,\varsigma,\varsigma',n,min,max$)}{$F$ is a set of source models, $\varsigma$ is a set of conditions associated with each source model, $\varsigma'$ is the target condition to be used for model generation, $n$ is the of data points per shape, $min$ and $max$ are $m$ by $r$ matrices (assuming $X_i \in \mathbb{R}^r$ and $m$ source models) with all minimum and maximum values, correspondingly.}
\BState \emph{Statistical Shape Model}:
\For {$i = 1 \rightarrow m$}
\State Define the input to sample from existing models: $X_i \gets GenerateInput(min_i,max_i,n)$
\State Get shape by merging inputs and output vectors: $S_i = [X_i, f_i(X_i)]$
\EndFor
\State Get the mean shape: $\bar{S} = \frac{1}{N} \sum_{i=1}^{N} S_i$
\State Get eigenvectors from PCA decomposition: $\phi \gets PCA(S)$
\State Get deformable parameters from PDM: $\boldsymbol{b} = \boldsymbol{\phi}^T ( S - \bar{S})$
\BState \emph{Hyper-Model}:
\State Train the hyper-model: $h :  b \rightarrow \varsigma$
\State Get the deformable parameter for new shape: $b' = h^{-1}(\varsigma')$
\State Get new shape: $S' = \bar{ \boldsymbol{S}} +  \boldsymbol{\phi}  b'$
\State Get input and output vectors from generated shape: $X', Y' \gets getInputOutput(S')$
\State Train a model for the new task. $f' : X' \rightarrow Y'$

\State \textbf{return} $f'$.

\EndProcedure
\end{algorithmic}
\end{algorithm}

Starting from the algorithm's arguments, the first difference is related with the $min$ and $max$ where in HPM2 these represent matrices of size $m \times r$, where $m$ is the number of source models and $r$ is the number of input features. These two matrices are a set of minimum and maximum values for each input per source models, so all the shapes can be generated according to their boundaries. As already explained, the main purpose of the algorithm is to include both input and output information for the ZSL problem. Therefore, a shape now is composed by both feature spaces (line 5). Furthermore, the algorithm remains the same until line 13, where a segregation of inputs and outputs should be made to train a new model in line 14.

\section{The Beta Distribution scenario}\label{beta_distribution}

The main goal of this section is to present a theoretical example for the use of HPM algorithm. For that, it would be ideal to have a set of simple functions with different properties, only with one input and output features so the algorithm could be well understood, and also with different observed output values for the same input value. On one hand, simple functions are suitable for this case because a visual feedback can be simply depicted, and on the other hand, with different types of functions the purpose of the HPM can be better grasped.

As the name of the section indicates, for this scenario we will be using a set of functions produced by the beta distribution. This distribution has two different parameters and by providing an input $x \in \mathbb{R}$, a different response $y \in \mathbb{R}$ is observed. By varying these parameters, the shape of functions will also vary. The expression for the probability distribution function (PDF) is as follows:

\begin{equation} 
f(x) = \frac{x^{\alpha-1}(1-x)^{\beta-1}}{B(\alpha,\beta)}
\end{equation}

where $x$ is the 1-dimensional input, $\alpha$ and $\beta$ are the parameters of the distribution and $B(\alpha,\beta)$ is defined as:

\begin{equation} 
B(\alpha,\beta) = \int_{0}^{1} t^{\alpha-1}(1-t)^{\beta-1}dt
\end{equation}

\begin{figure}[!ht]
\vskip 0.2in
\begin{center}
\centerline{\includegraphics[width=5.5in]{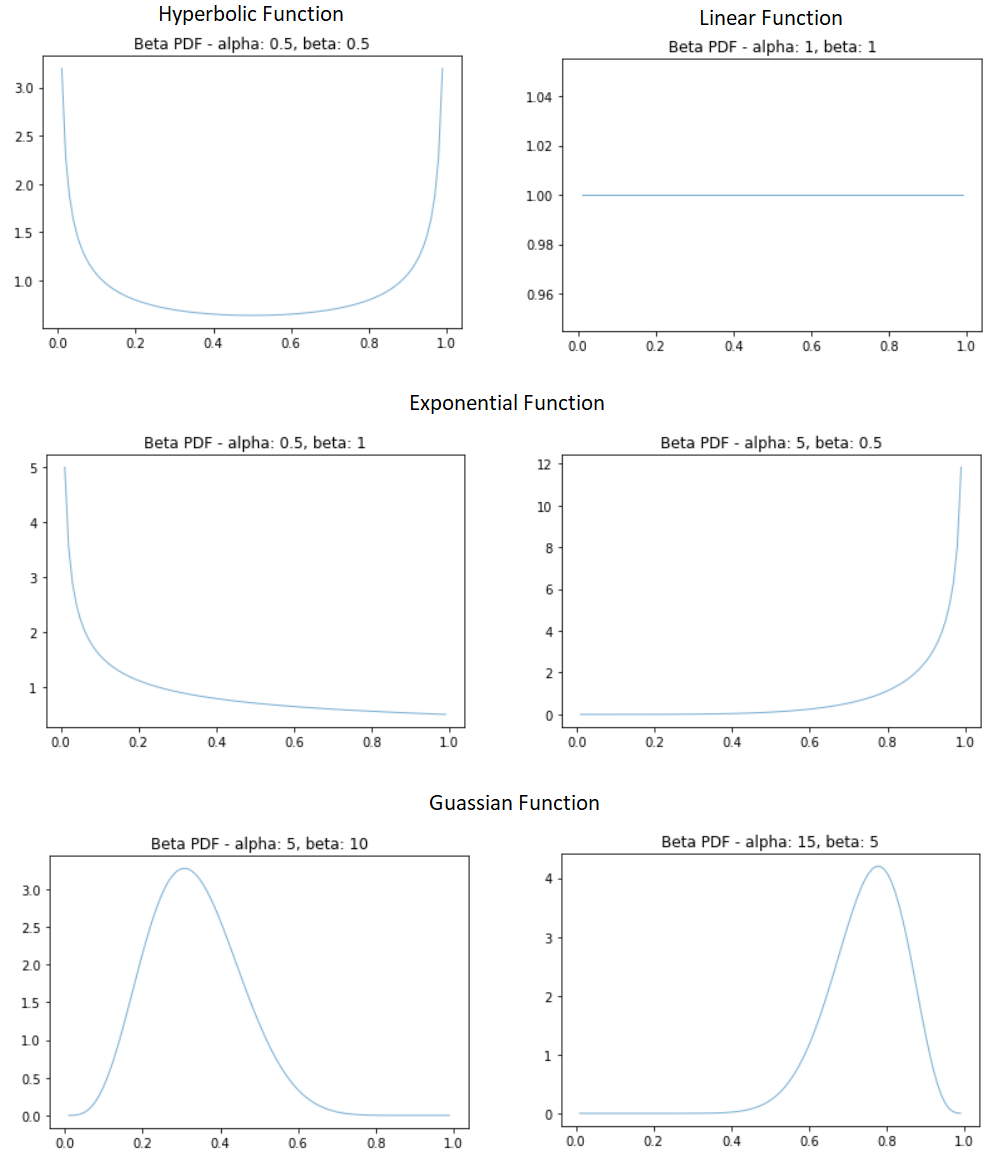}}
\caption{4 different types of functions generated by the beta distribution. In the top row, a hyperbolic function is depicted in the left handed side with $\alpha$ and $\beta$ values of [0.5,0.5] together with a linear function in the right handed side with $\alpha$ and $\beta$ values of [1,1]. In the mid row, two exponential functions are shown with $\alpha$ and $\beta$ values of [0.5,1] and [5,0.5]. In the bottom row, two Gaussian functions were generated using the $\alpha$ and $\beta$ values of [5,10] and [15,5].}
\label{beta_all}
\end{center}
\vskip -0.2in
\end{figure}

By using the beta distribution PDF it is possible to get a set of heterogeneous functions suitable to demonstrate a ZSL problem and apply the HPM to address it. For this case, 25 different functions were generated based on all combinations between $\alpha$ and $\beta$ values of 0.5, 1, 5, 10 and 15. These set of values were chosen because they generate a set of 4 different types of functions. These are depicted in Figure \ref{beta_all} where 6 different graphs are shown and grouped by hyperbolic and linear functions (top row), exponential functions (mid row) and Gaussian functions (bottom row). The goal for this example is to predict curves outside the 25 functions generated by the presented $\alpha$ and $\beta$ values. Therefore, in the training phase only the curves generated by the 25 set of beta distribution parameters will be used, where in the test phase 16 different combinations of $\alpha$ and $\beta$ parameters will be used in HPM algorithm to generate the corresponding curves. Here, the conditions for the hyper-model will be the $\alpha$ and $\beta$ parameters to train the hyper-model. The set of values for $\alpha$ and $\beta$ parameters in the test phase are 4, 6, 8 and 12. In order to generate the curves, 20 input values between 0.01 and 0.99 were used to ensure a fair representation of each curve. As for the machine learning techniques used, polynomial regression was chosen with multiple degrees for training both source models and hyper-model in HPM algorithm. In this case, cross-validation was not an option due to the limited number of existing data, specifically 25 for training and 16 for testing. Additionally, as one can see, the chosen test curves from beta distribution are not biased to achieve a better performance on HPM, where all the $\alpha$ and $\beta$ parameters lie between the parameters of training curves. Hence, we consider this experiment valid and fair to perform.

In order to clearly highlight the advantages of HPM in a simple way, this approach will be compared with the one presented by \cite{larochelle2008zero} and \cite{pollak2016models} where the coefficients of the base functions from the source models are used to train the model of models (hyper-model). As described earlier, we have stated the limitations of this approach saying that the modeling technique needs to be the same for all the source tasks, where different complexities in the datasets might exist. This means that certain techniques might be better than others for specific datasets. Hence, we should call the approach from \cite{pollak2016models} simply as hyper-model (HM).

\begin{figure}[!ht]
\vskip 0.2in
\begin{center}
\centerline{\includegraphics[width=5.5in]{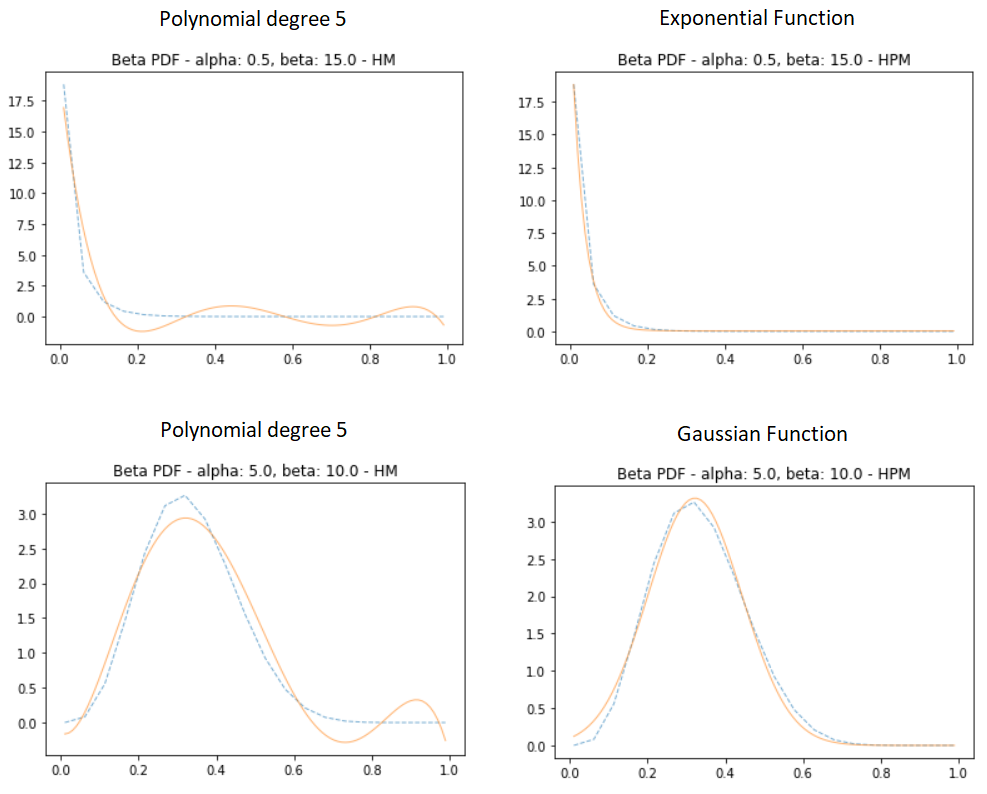}}
\caption{4 different graphs are shown, where the left sided graphs are the fits for Polynomial degree 5 and the right sided graphs are the curve fits for Exponential and Gaussian functions.}
\label{beta_hm_hpm}
\end{center}
\vskip -0.2in
\end{figure}

To highlight the differences between these approaches, Figure \ref{beta_hm_hpm} shows the fit of two curve types from beta distribution. The dashed blue lines are the training curves and the orange lines are the corresponding fits. On the left side are the fits from a polynomial degree 5 and on the right side the exponential and Gaussian curve fitting results. The main idea is to clearly see that a technique that tries to fit all curves have a lower performance than the most suitable curve fitting techniques for each task. As can be seen, on the left are the curves that the HM will use, and on the right side the shapes that the HPM will use to build the hyper-model.

At this point, in order to better differentiate between HM and HPM, we will briefly define its differences. For the HM, the models of models is trained by mapping model coefficients into task description / conditions. Therefore, a function $h : \lambda \rightarrow \varsigma$ should be trained, where $\lambda$ are the model coefficients from a single technique for all the models and $\varsigma$ the task description. Contrary to this, the HPM approach uses deformable parameters from a deformable model proposed by the SSM concept based on a set of shapes, where a shape corresponds to each model. Hence, a function $h : b \rightarrow \varsigma$ should be trained, where $b$ are the deformable parameters and $\varsigma$ the task description.

As stated before, the training phase for the hyper-model is composed by 25 parameters and task descriptions from the source tasks where machine learning techniques can be used to train models that fit these curves and are good generalizations. As in the HM approach, the same technique needs to be used for all tasks, so polynomial regression with varying degrees was chosen due to its flexibility to fit a large spectrum of curves. As for the HPM approach, exponential, Gaussian and polynomial degree 7 were used to model these curves. By analyzing the shape of the curves, it is possible to choose the best technique to fit the data, representing a clear advantage for HPM.

For both approaches, different types of hyper-parameters were tested. Regarding HM, different degrees for the polynomial regression were chosen to model the source tasks, while for HPM the best models were trained according to data shape and different number of components were tested when performing PCA. In both cases, the degree and number of components tested were 3, 4, 5 and 6. As for training the hyper-model in HPM and HM, polynomial regression was again used with degrees 3, 4, 5 and 6. For evaluation metrics, the mean of mean squared error (MSE) for each of the test cases was calculated, together with the standard deviation of the MSE. This represents a total of 32 tests performed, 16 for each approach. To test the generalization capabilities of each approach, the generated curves from each approach were tested against the ground truth from the beta distribution with each curve containing 100 datapoints. As previously explained, for the training process only 20 datapoints per curve were used to train the source models. This way we could assess the performance of the proposed approach in a more broad, robust and generic scenario beyond the 20 points when generating a new curve. For the HPM, 100 landmarks per shape were used taking advantage on the generalization capabilities of using the most suitable technique for each dataset according to its properties.

\begin{table}[t!]
\caption{HM approach: Mean MSE, standard deviation MSE and hyper-model $R^2$ for number of components from PCA decomposition of 3, 4, 5 and 6, and the degrees of 3, 4, 5 and 6 for polynomial regression of the hyper-Model.}
\label{beta_example_hm}
\vskip 0.15in
\begin{center}
\begin{small}
\begin{sc}
\begin{tabular}{ccccc}
\hline
Model Degree & \begin{tabular}[c]{@{}c@{}}Hyper Model\\ Degree\end{tabular} & \begin{tabular}[c]{@{}c@{}}Mean \\ MSE\end{tabular} & \begin{tabular}[c]{@{}c@{}}Standard Dev. \\ MSE\end{tabular} & \begin{tabular}[c]{@{}c@{}}Hyper-Model\\ $R^2$\end{tabular} \\
\hline
\textbf{3}   & \textbf{3}                                                   & \textbf{0.48}                                       & \textbf{0.116}                                               & \textbf{0.942}                                                             \\
4            & 3                                                            & 0.545                                               & 0.171                                                        & 0.782                                                                      \\
5            & 3                                                            & 0.52                                                & 0.185                                                        & 0.799                                                                      \\
6            & 3                                                            & 0.488                                               & 0.183                                                        & 0.703                                                                      \\
3            & 4                                                            & 0.613                                               & 0.381                                                        & 0.984                                                                      \\
4            & 4                                                            & 0.553                                               & 0.398                                                        & 0.968                                                                      \\
5            & 4                                                            & 0.652                                               & 0.288                                                        & 0.954                                                                      \\
6            & 4                                                            & 0.799                                               & 0.226                                                        & 0.913                                                                      \\
3            & 5                                                            & 0.971                                               & 0.711                                                        & 0.995                                                                      \\
4            & 5                                                            & 1.074                                               & 0.852                                                        & 0.973                                                                      \\
5            & 5                                                            & 1.258                                               & 0.882                                                        & 0.984                                                                      \\
6            & 5                                                            & 1.469                                               & 1.047                                                        & 0.971                                                                      \\
3            & 6                                                            & 2.676                                               & 2.394                                                        & 0.999                                                                      \\
4            & 6                                                            & 5.111                                               & 5.773                                                        & 0.997                                                                      \\
5            & 6                                                            & 6.367                                               & 6.321                                                        & 1                                                                          \\
6            & 6                                                            & 6.691                                               & 6.899                                                        & 0.999                                                                     
\\
\hline
\end{tabular}
\end{sc}
\end{small}
\end{center}
\vskip -0.1in
\end{table}

As presented in HPM and HPM2 the definition of the hyper-model states that it should be trained by mapping model coefficients into conditions / task descriptions. This was first formulated as such because normally the number of coefficients is greater than the number of conditions. However, as already discussed, it is a non trivial problem to find an inverse for some of the techniques used in machine learning. However, another way to handle such a problem is to use a search algorithm and find the optimal or near-optimal parameters that minimizes the distance between predicted and target conditions. As for this example, we would like to avoid both problems, and directly map conditions into model coefficients. Hence, multiple models were trained where each one maps all conditions into only one model coefficient. This implies that the number of models trained is the same as the output feature space dimension, which in this case are the number of source model coefficients of deformable parameters. This way, the hyper-model is a composition of multiple models, where each source model coefficient is predicted independently from the remaining. This should not be mistaken as ensemble regression, where multiple weak learners are trained with the same input and output features, and then the output of all learners are combined to produce a final prediction. 

As for the results of both approaches, Tables \ref{beta_example_hm} and \ref{beta_example_hpm} present all the 16 tests performed per approach, where in bold font the best result is depicted. In this case, the best result should be considered as the minimum value for mean MSE for all 16 tests performed in the test set. Complementarily to the evaluation metrics, also the coefficient of correlation $R^2$ for the hyper-models in HPM and HM is presented, so the reader can have a clear idea of its performance depending on the polynomial regression degree. Just before commenting the achieved results, one should highlight that for HM the number of models coefficients used to train the hyper-model is the value \textit{Model Degree} in Table \ref{beta_example_hm} plus 1 because of the bias term used for training. On the contrary, the \textit{Number of Components} in Table \ref{beta_example_hpm} is exactly the number of parameters used to train the hyper-model.

\begin{table}[t]
\caption{HPM approach: Mean MSE, standard deviation MSE and hyper-model $R^2$ for number of components from PCA decomposition of 3, 4, 5 and 6, and the degrees of 3, 4, 5 and 6 for polynomial regression of the hyper-model.}
\label{beta_example_hpm}
\vskip 0.15in
\begin{center}
\begin{small}
\begin{sc}
\begin{tabular}{ccccc}
\hline
Number Components & \begin{tabular}[c]{@{}c@{}}Hyper-Model\\ Degree\end{tabular} & \begin{tabular}[c]{@{}c@{}}Mean \\ MSE\end{tabular} & \begin{tabular}[c]{@{}c@{}}Standard Dev. \\ MSE\end{tabular} & \begin{tabular}[c]{@{}c@{}}Hyper-Model\\ $R^2$\end{tabular} \\

\hline
3                 & 3                  & 0.45                                                & 0.202                                                             & 0.92                                                                       \\
4                 & 3                  & 0.49                                                & 0.194                                                             & 0.818                                                                      \\
5                 & 3                  & 0.492                                               & 0.193                                                             & 0.683                                                                      \\
6                 & 3                  & 0.489                                               & 0.202                                                             & 0.615                                                                      \\
3                 & 4                  & 0.464                                               & 0.326                                                             & 0.964                                                                      \\
\textbf{4}     & \textbf{4}     & \textbf{0.32}                                     & \textbf{0.198}                                                 & \textbf{0.951}                                                              \\
5                 & 4                  & 0.45                                                & 0.181                                                             & 0.9                                                                        \\
6                 & 4                  & 0.638                                               & 0.193                                                             & 0.859                                                                      \\
3                 & 5                  & 0.677                                               & 0.6                                                               & 0.99                                                                       \\
4                 & 5                  & 0.601                                               & 0.467                                                             & 0.974                                                                      \\
5                 & 5                  & 0.838                                               & 0.61                                                              & 0.965                                                                      \\
6                 & 5                  & 1.173                                               & 0.896                                                             & 0.962                                                                      \\
3                 & 6                  & 2.208                                               & 1.88                                                              & 0.997                                                                      \\
4                 & 6                  & 3.473                                               & 3.313                                                             & 0.996                                                                      \\
5                 & 6                  & 5.127                                               & 4.672                                                             & 0.996                                                                      \\
6                 & 6                  & 6.276                                               & 6.323                                                             & 0.994                                                                     \\
\hline
\end{tabular}
\end{sc}
\end{small}
\end{center}
\vskip -0.1in
\end{table}

The first thing to refer is that the best set of parameters for HPM performs better than the ones in HM, with a mean MSE of 0.32 for 4 components and hyper-model polynomial degree 4, and 0.48 for model polynomial degree 3 and hyper-model polynomial degree 3 (which is very similar to mean MSE of model degree 6 and hyper-model degree of 3), correspondingly. This supports our hypothesis that by using the SSM concept that can take advantage on different techniques by dealing with shapes instead of model coefficients, a better performance can be achieved when comparing with HM. The second aspect to notice is that by making a direct comparison between each of the tests from both approaches where the model degree and number of components are the same together with the hyper-model degree (we might call it as the same setting), for all tests the HPM performs better. This means that the HPM is more effective than HM for the same problem. Additionally, as previously explained, the number of model coefficients from the polynomial in HM to train the source models corresponds to the model degree plus 1, so for the same setting, the HPM is more efficient because it uses a lower dimension for the problem.

\begin{figure}[!ht]
\vskip 0.2in
\begin{center}
\centerline{\includegraphics[width=5in]{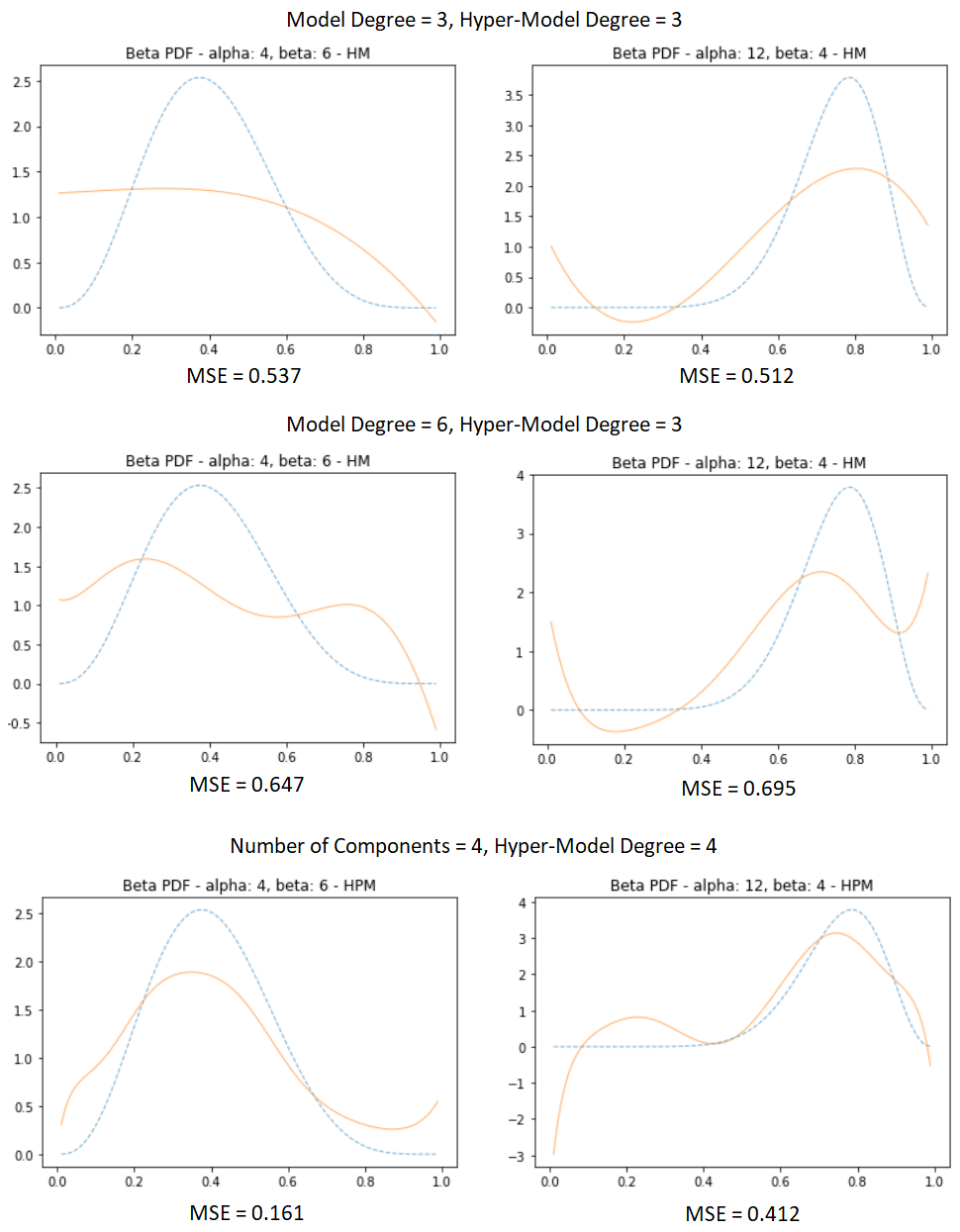}}
\caption{Graph plots of two test curves, where the dashed blue line is the ground truth and the orange line is the result of the ZSL approach applied. In the left column are the results for beta Distribution with $\alpha$ = 4 and $\beta$ = 6, and in the right column the results for  $\alpha$ = 12 and $\beta$ = 4. The first row corresponds to the HM approach with model degree of 3 and hyper-model degree of 3, to mid row refers also to HM approach but with model degree 6 and same hyper-model degree, and the bottom row is the HPM with 4 components and hyper-model degree of 4.}
\label{hm_hpm_results}
\end{center}
\vskip -0.2in
\end{figure}

For a better interpretation of the above results, Figure \ref{hm_hpm_results} presents two test scenarios for the beta distribution with $\alpha$ = 4 and $\beta$ = 6 on the left and $\alpha$ = 12 and $\beta$ = 4 on the right, where the dashed blue line is the ground truth from beta distribution and the orange line is the result of the ZSL approach applied. In each row, different settings were used: On the top, HM was used with model degree of 3 and hyper-model degree of 3; On the mid row, again the HM approach was used with model degree of 6 and hyper-model degree of 3; On the bottom row, the HPM approach was tested with 4 components and hyper-model degree of 4. Additionally, below each image the MSE is depicted to assess the distance between the predicted curve and the ground truth. As can be seen, the best results are yielded by the HPM approach and, for the HM, it can also be seen that the MSE increased as the model degree increased from 3 to 6. It can be clearly seen the effect of increasing the polynomial degree used to fit the source models as in the mid row the predicted curves are more complex and irregular than the ones in the top row. Despite being more complex, they do not produce better results and are not a better suit for the presented ground truth.

Hence, for the beta distribution scenario we have shown that HPM is, first, more effective than HM due to better performance in all the corresponding settings / tests performed, and second, more efficient because it addresses the same problem with less complexity than the HM approach. This is what we were referring previously as by taking advantage on the regression setting, the performance of ZSL techniques could be improved.

\section{Discussion and Main Conclusions}\label{discussion_concluion}



As a main motivation to create new techniques for the machine learning community, specifically for the ZSL area, and consequently building technologies that allow, e.g. to assist the integration of new product parts or new machines in an industrial scenario, the HPM algorithm is proposed. From the review of current state of the art techniques to its assessment in test scenarios, the main goal of developing new technology is always assisting humans to perform a more effective and efficient job, or even replace them in case of high risk situations avoiding catastrophic repercussions. 


%
%

As already described in Section \ref{chap_hyper_model} when presenting the HPM algorithm, one of the advantages is the shape analysis it performs to the data itself using a deformable model. By analyzing how data varies from model to model, an unsupervised feature space that represents the main modes of deformation was defined and used to generate back new shapes based on new process conditions. Comparing with the hyper-model (HM) approach, using data directly allows for a more detailed analysis of system dynamics. Despite the model coefficients are also related with data shape, in certain situations the HM approach can overfit or underfit the data. Since the main restriction of HM is the assumption that a "model-fits-all" datasets, the same technique will have different performances according to the dataset in hands, and can ultimately overfit data. However, for fairness of comparison, we should highlight that the effect of overfitting can be managed by the use of regularizers, such as LASSO or ridge regression. Since the regularizers allow to shrink the coefficient values to decrease model variance, by using the well known cross-validation strategy the penalization parameters could be optimized in order to find the best model coefficients. Regarding the case of underfitting, this can also be avoided by not using too simple models, e.g. by increasing the polynomial degree. Since regularizers can be used to tackle model complexity, a fair model can be trained that does not suffer too much from both overfitting and underfitting effects. 

Although this seems as a rule of thumb that can be applied to the HM, one should be aware that a compromise between underfitting and complexity should be made. Let's image that in an ideal situation a relatively high polynomial degree is used to avoid underfitting, and the use of regularizers will address the overfitting effect. In this case, since the polynomial degree is high, so is the number of parameters that the hyper-model in HM should handle, and more complex is the problem. In this case, domain drift can occur, where the trained source models are near optimal, but the hyper-model might not be due to high number of coefficients from the source models. Hence, the solution to this problem is to lower the number of parameters in the polynomial regression when training source models. Now the complexity that the hyper-model in the HM approach should handle is more acceptable and we can state that an optimal hyper-model can be trained. However, since the polynomial degree is decreased once training the source models, in certain situations, the models can be too simple to grasp the system details and generality is lost. This is the case where domain drift occurs again, but now with the source models being sub-optimal and the hyper-model optimal. Hence, a good trade-off between source models and hyper-model complexity should be achieved.

In case of HPM, since a shape analysis is made using a decomposition method, the number of eigenvectors that map the shape back to its original feature space from a set of deformable parameters should also be chosen. Hence, this trade-off also exists for HPM. At this point, both approaches are side-by-side. The main difference starts when the learning process of HPM is made jointly with all source tasks available, while for the HM each set of coefficients should be learned for each task separately. Therefore, a holistic perspective is taken for HPM and a more local one for the HM. On top of all that, one should not forget that the most suitable technique for each dataset can be chosen in the HPM. This is an advantage because if the model is a good generalization of data, the predictions made by the model can be considered reliable to be used for the shape analysis of HPM. If these models are not good generalizations, which might occur in HM, valuable information is not grasped and the hyper-model will be trained with a limited view of the system dynamics.

In the presented scenario using the beta distribution the results show the advantage of using a shape-based technique rather than using source model coefficients directly to extrapolate task relations in a regression setting. As previously discussed, in the presented scenario the benefits of HPM are depicted, but we believe that a more complex scenario would make clear the advantage of HPM, and in concrete, the shape analysis performed on data. This more complex scenario could be easily imagined where source tasks represent non-linear systems and ANNs might be used to train those tasks. Normally, an ANN for a regression setting has multiple hidden layers and neurons per hidden layer, being composed by tens, hundreds or even thousands of weights to optimize. In that sense, we consider the use of ANNs to model non-linear systems a more likely scenario to happen in real world machine learning applications. Therefore, the use of HM will be very limited because one should train a hyper-model using all the weights from multiple source models. From this, we have a too high dimension to deal with and is impractical to train a hyper-model with such high dimension in the input feature space. Additionally, since the hyper-model maps model coefficients into conditions, if new conditions are given to generate a new model, the problem of finding the most suitable model coefficients would be too complex. Additionally, if one consider the matching between all source model ANNs weights, also the HM might not be the most suitable technique. As already explored thoroughly in literature, two slight different ANNs, particularly the multi-layer perceptrons (MLPs), might provide very similar results but with very different weight values due to the stochastic learning routines that can be used. Although the same thing does not happen in polynomials where similar models normally have close coefficients, these might not be capable of modeling some non-linear systems.

Opposite to this, the HPM does not suffer from such issues. Even if the source models are all ANNs with thousands of weights, since a shape analysis is made on the data itself in HPM the dimensionality is greatly reduced and a hyper-model can be trained. Moreover, since HPM does not assume any model coefficient matching from different source models, it does not have any problem dealing with machine learning techniques with a high number of coefficients to optimize, as far it maximizes generality. This way, by only hypothesizing about a more complex scenario where ANNs are used, one can easily understand the inherent benefits of HPM.

Finally, we would like to bring into discussion an additional potential problem that HPM can address. This kind of problem is not related with how to better learn a new problem of interest, as the definition of ZSL stated, but which problem of interest is worth pursuing from a large spectrum of possibilities. The intuition behind such an exploratory approach using ZSL for regression if based on the possibility to generate new models for a continuum of conditions. Based on this, a vast number of models can be efficiently generated from a pool of source models, and if correctly assessed, exclude the ones not worth to investigate and select the most promising ones.

For this case, lets think about a new scenario. Imagine that a pharmaceutical company wants to explore new processes, like the combination of different chemical agents that have different interactions among them to build new drugs. In a pool of chemical agents, the amount of possible combinations is very high and it is not feasible to try all these combinations. This kind of approach of trying all combinations is similar to "exploration" in search algorithms that need to search for all solution space to find the optimal solution. Rather, a more exploitation-based approach can be taken. In this case, if information about the chemical agents (e.g. some chemical properties as task descriptions), the end result of their reaction and the parameters used for their combination (e.g. properties about the medium used for the reaction), models can be trained to predict the outcome of a reaction based on the experimental parameters, for a specific combination of agents. Hence, by applying the HPM algorithm, models can be generated by providing the new target combination of agents. From this point, a pool of new models can be generated using HPM and ranked according to drug specifications. Then, the best combination of chemical agents used to generate the models can be tested in the lab to provide a more oriented search for the final drug result. Ultimately, if new models are trained based on these new lab experiments, the pool of source models used by HPM can be enlarged. Hence, the process can be repeated until the optimal, or near-optimal, chemical agent combination is found. After experimentation, if the drug requirements are not yet met, a new iteration can be made, where the HPM can be retrained and generate a new set of candidate predictive models. This iterative approach can be performed as many times as required, and in principle, it should be more efficient than experimenting all possible combinations of chemical agents. Even if the experimental results are very different from the model generations using HPM in the first iteration, since the HPM is updated in every iteration with new models, it becomes better and better at generating models and will eventually start converging. The intuition behind this process, is to basically create a gradient among models so that trying all the agent combinations in the lab can be avoided. 

Of course humans do not perform these kind of experiments blindly and choose the next chemical agents to try at the lab randomly, but rather based on knowledge about the field and previous performed experiments. This last presented approach is performed by humans in almost every task that has a clear goal to fulfill. By measuring how far the result is from the intended goal, fewer changes are tried and better results could be achieved. This is applicable as well in our example, where humans can collect knowledge about the chemical agents to experiment based on a set of previous experiments. However, this gets more challenging when the problem in hands is far too complex in order for humans to understand the gradient-progressing towards the final goal, and often the selection of new experiments is based on intuition. When there is a high number of variables describing the experiments and its end results, machine learning algorithms can help to make a more informative approach in such complex problems, avoiding to use intuition that is hard to justify and no knowledge can be exploited from that. Hence, the main consideration to get is not to blindly rely on machine learning algorithms, mainly because it is not possible to quantify and translate all human knowledge to an algorithm, and neither to completely discard these. Opposite to this, these ZSL algorithms can be seen as complementary tools that humans can take advantage from and ease the burden of hard and heavy work that sometimes one needs to perform.




\vskip 0.2in
\bibliography{myrefs}

\end{document}